\pdfoutput=1

\documentclass[11pt]{article}

\usepackage[]{acl}
\usepackage{comment}
\usepackage{tikz}
\usepackage{times}
\usepackage{latexsym}

\usepackage[T1]{fontenc}

\usepackage[utf8]{inputenc}

\usepackage{microtype}
\usepackage{graphicx}
\usepackage{tabularx}
\usepackage{xspace}
\usepackage{amsmath}

\newcommand{\camerareadytext}[1]{\xspace}

\title{Sociodemographic Prompting is Not Yet an Effective Approach for Simulating Subjective Judgments with LLMs}

\author{
  Huaman Sun \\
  University of Toronto \\
  \texttt{hm.sun@mail.utoronto.ca}
  \And
  Jiaxin Pei \\
  Stanford University \\
  \texttt{pedropei@stanford.edu}
  \AND
  Minje Choi \\
  Amazon \\
  \texttt{minjec@amazon.com}
  \And
  David Jurgens \\
  University of Michigan \\
  \texttt{jurgens@umich.edu}
}

\begin{document}
\maketitle

\begin{abstract}

Human judgments are inherently subjective and are actively affected by personal traits such as gender and ethnicity. While Large Language Models (LLMs) are widely used to simulate human responses across diverse contexts, their ability to account for demographic differences in subjective tasks remains uncertain.
In this study, leveraging the \textsc{Popquorn} dataset, we evaluate nine popular LLMs on their ability to understand demographic differences in two subjective judgment tasks: politeness and offensiveness. We find that in zero-shot settings, most models' predictions for both tasks align more closely with labels from White participants than those from Asian or Black participants, while only a minor gender bias favoring women appears in the politeness task. Furthermore, sociodemographic prompting does not consistently improve and, in some cases, worsens LLMs’ ability to perceive language from specific sub-populations. These findings highlight potential demographic biases in LLMs when performing subjective judgment tasks and underscore the limitations of sociodemographic prompting as a strategy to achieve pluralistic alignment. Code and data are available at: \url{https://github.com/Jiaxin-Pei/LLM-as-Subjective-Judge}.

\end{abstract}

\section{Introduction}

From sentiment analysis to dialogue generation, large language models (LLMs) have demonstrated impressive capabilities in various natural language processing (NLP) tasks \citep{brown2020language, radford2019language}. Recent research has begun exploring whether these models possess social knowledge analogous to that of humans \citep{zhou2023sotopia, choi2023llms}. For example, \citet{almeida2024exploring} replicate eight classic psychological experiments on LLMs to test their ability to reason about moral and legal issues. \citet{yildirim2024task} examines how LLMs’ ``instrumental knowledge'' relates to the more ordinary "worldly" knowledge of human agents. Building on these insights, LLMs have been applied to large-scale labeling tasks requiring social understanding, and often with promising results \citep{ziems2023large, rytting2023coding}. In terms of subjective tasks, researchers have explored LLMs' zero-shot potential in areas such as character simulation \citep{wang2023rolellm} and hate speech detection \citep{plaza-del-arco2023-respectful}.

However, LLMs face significant challenges in handling subjective tasks. It is well acknowledged that social biases and stereotypes embedded in their training data can lead to inadequate representation of diverse human experiences \citep{santurkar2023opinions}. As a result, using LLMs for subjective tasks risks producing outcomes that disproportionately favor certain demographic groups, leading to biased or unfair results \citep{liang2021towards}. \citet{santurkar2023opinions} found that when responding to value-based questions, LLMs tend to align more closely with the perspectives of lower-income, moderate, and Protestant or Roman Catholic individuals. Despite these early findings, limited research has explored whether LLMs exhibit similar systemic biases with certain social groups across other subjective NLP tasks, highlighting the need for further investigation into their broader implications.

Subjective tasks present an additional challenge because language perception is shaped by social context and identity \citep{alkuwatly2020identifying}. For instance, a text perceived as polite or inoffensive by one group may be interpreted differently by another. Ideally, LLMs should capture the full spectrum of subjective judgments. Steerable pluralism, as described by \citet{sorensen2024roadmap}, refers to an LLM’s ability to be faithfully adjusted to represent specific perspectives. Yet, \citet{miehling2024evaluating} found that many current LLMs have limited steerability to take on various persona, due to both inherent biases in their baseline behavior and asymmetries in how they adapt across different persona dimensions. These limitations suggest that while steerability is a promising direction, it requires more refinement to effectively capture diverse perspectives.

Sociodemographic prompting, which involves enriching prompts with demographic or individual-specific information, has gained increasing attention in recent research. This approach has shown potential for improving data augmentation and simulating human behavior for social science applications \citep{hwang2023aligning, argyle2023out}. Despite its promise, the effectiveness of sociodemographic prompting remains debated, as model performance can be sensitive to the phrasing, structure, or order of prompts \citep{mu2023navigating, dominguez2023questioning}. For example, \citet{beck2024sensitivity} finds that the impact of adding demographic information varies significantly depending on the model, task, and prompt design. Moreover, some studies suggest that sociodemographic prompting can exacerbate stereotypes and biases \citep{deshpande2023toxicity} or reduce model performance on certain tasks \citep{santurkar2023whose}. 

Given these mixed findings and the focus of previous studies on specific NLP tasks, our work extends the literature by examining (1) whether LLMs' predictions systematically align more with certain social groups on two more subjective tasks and (2) how LLMs can effectively account for identity-based differences in perception when handling subjective language tasks with sociodemographic prompting. Leveraging the \textsc{Popquorn} dataset \citep{pei2023annotator}, we evaluate nine popular LLMs on their ability to understand demographic differences in subjective tasks, offensiveness and politeness. The two tasks are occasionally related but distinct. Politeness pertains to notions of status differences and interpersonal distance, while offensiveness involves violations of expected social norms. Offensiveness is not as broad as impoliteness, as varying levels of politeness can be perceived as non-offensive. Exploring these subtly different tasks offers a more comprehensive evaluation of LLMs’ potential biases in subjective NLP tasks.

Overall, our results reveal that intrinsic biases persist in LLMs when applied to these tasks. The study highlights the limitations of LLMs in understanding and aligning gender and racial differences in subjective judgment. While some research aims to directly use LLMs to simulate group-specific social behaviors, our findings underscore the risks of unintentionally reinforcing racial and gender biases when applying sociodemographic prompting to subjective tasks.

\section{Methods}

\paragraph{Data}
We use the \textsc{Popquorn} dataset~\cite{pei2023annotator} to evaluate LLMs' capacity to tackle subjective NLP tasks. \textsc{Popquorn} includes 45,000 annotations from a demographically representative U.S. sample. We focus our analysis on two identity types: gender and ethnicity. To ensure statistical robustness, we focus on the gender categories \texttt{Man}, \texttt{Woman} and ethnic groups \texttt{Asian}, \texttt{Black} and \texttt{White} as they have sufficient annotations.

For this study, we analyze annotators' offensiveness and politeness ratings on a 5-point Likert scale\camerareadytext{ (1 to 5)}. We compute average scores for each identity group to capture perceptions from specific demographics. The mean overall offensiveness score is 1.88 (SD = 0.76), and politeness scores average 3.31 (SD = 0.91). Scores from men, women, and White annotators closely mirror the overall distribution, while Black and Asian annotators show diverging means and higher variance. Figure~\ref{fig:score} in Appendix~\ref{appendix:data} shows the distributions of both overall and identity-specific scores for offensiveness and politeness tasks.

\paragraph{Models}
To enhance the generalizability of our findings, we conduct experiments with a range of open-source and close-source LLMs: FLAN-T5-XXL \citep{chung2022scaling}, FLAN-UL2 \citep{tay2023ul2}, Tulu2-DPO-7B, Tulu2-DPO-13B \citep{ivison2023camels}, GPT-3.5\camerareadytext{\footnote{https://platform.openai.com/docs/models/gpt-3-5}}, GPT-4 \citep{openai2023gpt4}, Llama-3.1-8B-Instruct \citep{dubey2024llama3}, Mistral-7B-Instruct-v0.3 \citep{jiang2023mistral7b}, and Qwen2.5-7B-Instruct \citep{qwen2025qwen25technicalreport}.

\paragraph{Prompts}
We design prompts to instruct the models to predict offensiveness and politeness scores for each instance. To ensure the prompts elicit valid responses, we conduct preliminary experiments on a small subset of data. An example prompt (Table~\ref{tab:prompt}) and the full list of prompts used in our study (Table~\ref{tab:prompt_list}) are shown in Appendix~\ref{appendix:prompt}.
We test the robustness of our results using different prompt templates and option orders (i.e., 1 to 5 or 5 to 1) on a set of open-source LLMs. Overall, we observe minor differences in LLMs' performance across templates and option orders. Details are provided in Appendix~\ref{appendix:prompt}.

\begin{figure*}[t!]
\centering
    \includegraphics[width=1.0\textwidth]{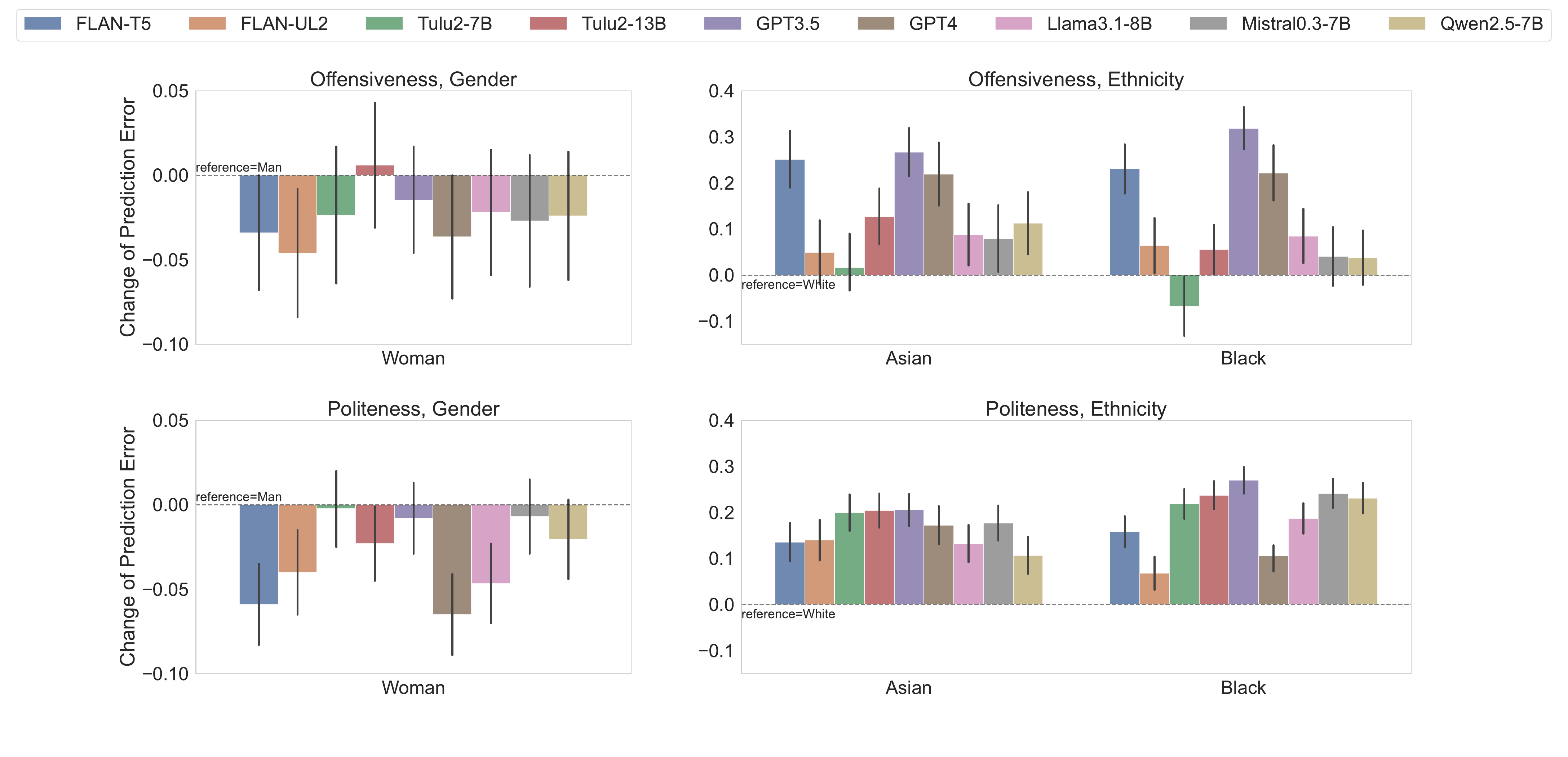}
    \caption{Regression results for predicting the gap between model predictions and the labels from each demographic group. The models’ predictions for offensiveness are not significantly different from the ratings by Men and Women except for FLAN-UL2 (Top left). However, LLMs’ predictions are significantly closer to Women’s ratings for politeness (Bottom left) and are closer to White people’s ratings compared with ratings from Black and Asian annotators in both tasks (Right).}
    \label{fig:base}
\end{figure*}

\section{Are Model Predictions Closer to Certain Demographic Groups?}

While individual judgments may vary, LLMs can generate only a single prediction unless explicitly instructed to output a distribution. Therefore, when LLMs are applied to judgment tasks, it is crucial to examine whether their predictions align more closely with certain demographic groups.

\paragraph{Analysis} To measure the alignment between LLM and certain demographic groups, we define baseline prediction error (\(E_{base}\)) as the absolute difference between LLMs’ predictions using identity-free prompts and human ratings from a specific demographic group:
\[ E_{base} = | \text{prediction} - \text{label}_{subgroup} | \]
For each task and demographic identity type, we apply separate linear mixed effect models to examine changes in baseline prediction error of a specific demographic group (target group) compared to the reference group, controlling for instance-level variations with instance ID as a random effect. For example:
\[ E_{base} = \beta \text{gender} (\text{ref}=\text{man}) + (1|\text{instance id}) \]
A regression coefficient \( \beta = 0 \) indicates that there is no difference in baseline prediction errors between the target and reference groups. A positive \( \beta \) means that baseline prediction errors are larger for the target group, suggesting that the LLM predictions are closer to the reference group than to the target group. The aggregated results are visualized in Figure~\ref{fig:base}, while Table~\ref{tab:basegap} in Appendix~\ref{appendix:results} provides detailed results from the linear mixed effects regressions.

\paragraph{Results} As shown in Figure~\ref{fig:base}, LLMs' baseline prediction errors for offensiveness do not show significant gender differences, except for FLAN-UL2. This is expected as the original \textsc{Popquorn} paper \citep{pei2023annotator} reports no significant gender differences in human ratings of offensiveness. However, for politeness ratings, LLM predictions tend to align more closely with women's ratings, except for GPT-3.5 and Tulu2-7B. Surprisingly, for both Tulu2 and GPT models, those with more parameters exhibit a greater bias in politeness prediction, suggesting that simply scaling models may not effectively reduce biases in subjective tasks. Furthermore, LLMs' predictions for both politeness and offensiveness are consistently closer to the ratings of White annotators compared to those of Black or Asian annotators. This result reflects the intrinsic bias of LLMs on subjective judgment tasks.


\begin{figure*}[t!]
\centering
    \includegraphics[width=0.95\textwidth]{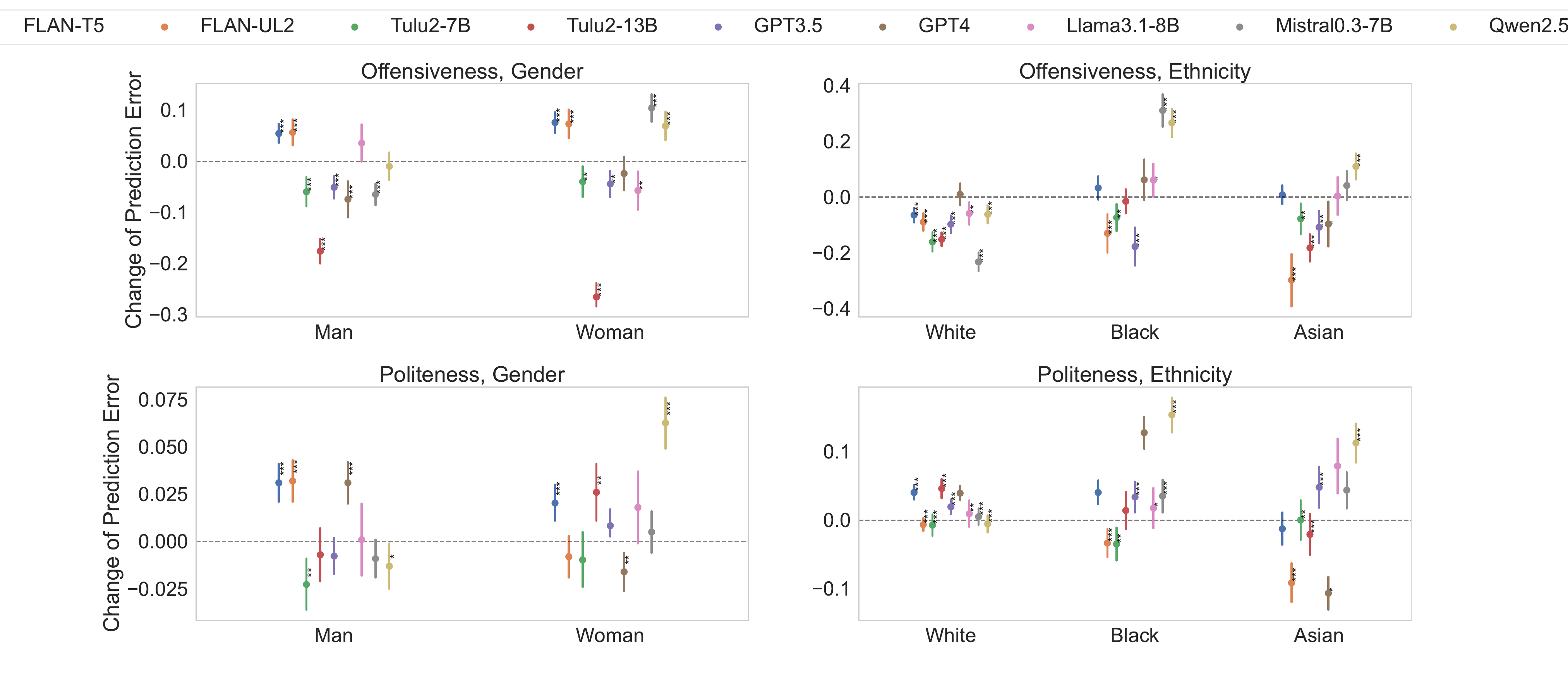}
    \caption{Regression results for predicting the prediction errors with different prompt settings. Each point shows the change of prediction errors when adding identity to the prompt for both tasks, relative to an identity-free prompt. Overall adding demographic tokens in prompts does not consistently improve the LLMs' performance for predicting ratings from different demographic groups. }
    \label{fig:gap}
\end{figure*}

\section{Does Sociodemographic Prompting Improve Alignment with Demographic Groups?}

Given the intrinsic bias of LLMs in subjective judgment tasks, a natural question arises: does adding demographic information in prompts steer LLMs to generate more diverse outputs that better align with specific groups? In this section, we conduct a series of analyses to answer this question.

\paragraph{Analysis} To tackle this research question, we modify the prompt in Appendix~\ref{appendix:prompt} Table~\ref{tab:prompt} and add demographic information when prompting the model to predict group-based ratings on offensiveness and politeness (e.g., ``How offensive does a White person think the following text is?''). 
We then further run separate linear mixed-effect regression models to predict the change in the model's absolute prediction errors when being prompted with and without demographic tokens. Instance IDs are controlled as a random effect to account for the instance level variations.
Figure~\ref{fig:gap} illustrates the change in model performance when adding identity tokens into prompts. In the plots, points above 0 indicate that incorporating an identity token increases the model's prediction errors, while points below 0 suggest that the identity token improves prediction performance. Detailed regression results are provided in Table~\ref{tab:addgap}, Appendix~\ref{appendix:results}.

\paragraph{Results} In Figure~\ref{fig:gap}, our analysis reveals that in certain cases, identity tokens help models adjust their predictions. For instance, adding an ethnicity token improves GPT-3.5 and FLAN-UL2’s ability to predict offensiveness ratings from Asian participants. However, this improvement is not consistent across tasks and models. While adding an ethnicity token helps GPT-3.5 better predict offensiveness ratings from Black participants, it has no effect on GPT-4. In contrast, identity tokens actually increase prediction errors for politeness ratings from Black participants in both GPT-3.5 and GPT-4. These findings highlight the challenges of mitigating LLM prediction biases in subjective NLP tasks and suggest that incorporating sociodemographic information in prompts is not yet a reliably effective solution.

\section{Discussion}
With the large-scale deployment of LLMs in our society, it becomes increasingly important to study whether LLMs are able to understand the preferences of different groups of people. Our results suggest that LLMs are more aligned toward certain demographic groups than others on subjective perception tasks. For both of our tasks, we find that all of our tested LLMs provide answers which are closer to the annotations of White annotators compared to other demographic groups. Our findings contribute to the newly growing knowledge of types of demographic biases inherent in LLMs when asked to solve subjective tasks~\cite{feng2023pretraining}, signaling caution for potential applications such as deploying LLMs for generating annotations at large scale~\cite{ziems2023large}.

Our results also suggest that directly inserting demographic features into prompts, unfortunately, does not reliably help models adopt the perspectives of target groups. 
The ability of LLMs to consider various opinions, at least from the perspective of demographic groups, seems limited at its current stage. Furthermore, we observe that newer models, such as Mistral-0.3 and Qwen-2.5, exhibit reduced alignment on different task types and identity-based prompts.  This may be due in part to increasingly strict guardrails designed to mitigate harmful outputs, which can also affect model performance by increasing refusal rates and limiting functionality \citep{bonaldi2024safer}. Given that our tasks include sensitive keywords (e.g., \textit{vulnerable identity, offensive, not polite}), these safety mechanisms may further contribute to the diminished effectiveness of identity-based prompting in newer models.

\section{Conclusion}
In this study, we study LLMs capability to account for demographic differences in subjective judgment tasks. We find that LLMs' predictions are closer to White people's perceptions for both tasks and across 9 models compared with Asian and Black people. We further explore whether incorporating demographic information into the prompt helps mitigate this bias. Surprisingly, we find that adding identity tokens (e.g. \texttt{Black} and \texttt{Man}) does not consistently help to improve the models' performance at predicting demographic-specific ratings. 
Our results suggest that LLMs may hold implicit biases on subjective NLP tasks and sociodemographic prompting is not an effective approach to address this bias yet. Researchers and practitioners should be careful when using LLM as judges on subjective tasks.


\section{Ethics}
This study investigates LLMs' capability to represent the opinions of different demographic groups when producing answers for subjective NLP tasks such as detecting offensiveness and politeness. As LLMs are increasingly being deployed in various settings that require subjective opinions, the fact that their opinions are significantly biased towards certain gender and ethnic groups raises a problem in their ability to remain neutral and objective regarding different tasks. Especially, prior work has shown that LLMs can produce biased and toxic responses when generating text provided the personas of specific individuals \cite{deshpande2023toxicity}. When conducting studies on LLMs to understand how they can simulate the opinions or perspectives of a particular individual or social group, the research should be guided toward a direction that can overcome existing problems instead of introducing new problems such as AI-generated impersonation. Following, we discuss the ethical implications of our study.

During this study, we made the decision to only use the men and women gender labels from \textsc{Popquorn}, which unfortunately gives the appearance of an implicit binary assumption of gender. This choice is solely motivated by the absence of other gender identities in that dataset; while \textsc{Popquorn} is the largest and most diverse, due to the relative rareness of other gender identities in the crowdsourcing pool they used, no additional identities are available without additional data collection on our part, which we view as outside the scope of this paper. However, we acknowledge that our experiment settings miss out on non-binary forms of gender representation, which was inevitable due to data availability and how the original dataset was constructed. Nevertheless, the representativeness of non-binary individuals and groups in LLMs is also an important topic regarding potential disproportionateness. We call for future work in this direction to expand the inclusiveness of all types of social groups in their data collection.

When conducting large-scale analyses on datasets using LLMs, another topic of interest is minimizing financial costs and environmental impact. In this study, we do not require any finetuning or training stages and experiment only by inferring prediction results from publicly available LLMs. Except for GPT-3.5 and GPT-4, all models were able to run on a single A5000 GPU and took around six hours to run on the entire dataset under a single setting. 

\section{Limitations}
Our study has the following limitations: (1) Although we aim to include most updated and popular LLMs into the analysis, we only experiment with a limited number of them due to the computational cost of running these experiments. We will release all the scripts to allow future researchers to test other models' performance in understanding group differences. (2) In our experiment settings, we only select limited types of ethnicity and gender categories for analysis due to the sparsity of labels from people with other identities in the \textsc{Popquorn} dataset; therefore, our study didn't include several important identity groups such as non-binary genders and Hispanic people. (3) We only studied two tasks: offensiveness ratings and politeness ratings. As the datasets used for annotating these tasks come from offensive Reddit comments and polite emails, the biases reported in this study may not generalize to other datasets and task settings. (4) Our model predictions take the form of ordinal values, whereas the averaged annotation scores are fractional values. (5) We do not examine intersectional identities due to sparsity when subsetting the data, while the bias associated with populations defined by multiple categories leads to an incomplete measurement of social biases \citep{hancock2007multiplication}. (6) We observe that some models, particularly GPT3.5 and Tulu2, have a relatively high refusal rate when asked to providing ratings, especially for offensiveness task and when prompts involve specific demographic groups such as Black people. Table~\ref{tab:invalid_offensive} and Table~\ref{tab:invalid_polite} in Appendix~\ref{appendix:guardrail} present the percentages of invalid responses by models and identity prompts. These implicit guardrails of LLMs may affect our findings, as the models might recognize the context but decline to respond due to privacy or ethical concerns.

\bibliographystyle{acl_natbib}
\bibliography{anthology,custom}

\newcommand{\noop}[1]{}
\begin{thebibliography}{33}
\expandafter\ifx\csname natexlab\endcsname\relax\def\natexlab#1{#1}\fi

\bibitem[{Al~Kuwatly et~al.(2020)Al~Kuwatly, Wich, and
  Groh}]{alkuwatly2020identifying}
Hala Al~Kuwatly, Maximilian Wich, and Georg Groh. 2020.
\newblock \href {https://doi.org/10.18653/v1/2020.alw-1.21} {Identifying and
  measuring annotator bias based on annotators{'} demographic characteristics}.
\newblock In \emph{Proceedings of the Fourth Workshop on Online Abuse and
  Harms}, pages 184--190, Online. Association for Computational Linguistics.

\bibitem[{Almeida et~al.(2024)Almeida, Nunes, Engelmann, Wiegmann, and
  de~Ara{\'u}jo}]{almeida2024exploring}
Guilherme~FCF Almeida, Jos{\'e}~Luiz Nunes, Neele Engelmann, Alex Wiegmann, and
  Marcelo de~Ara{\'u}jo. 2024.
\newblock Exploring the psychology of llms’ moral and legal reasoning.
\newblock \emph{Artificial Intelligence}, 333:104145.

\bibitem[{Argyle et~al.(2023)Argyle, Busby, Fulda, Gubler, Rytting, and
  Wingate}]{argyle2023out}
Lisa~P Argyle, Ethan~C Busby, Nancy Fulda, Joshua~R Gubler, Christopher
  Rytting, and David Wingate. 2023.
\newblock Out of one, many: Using language models to simulate human samples.
\newblock \emph{Political Analysis}, 31(3):337--351.

\bibitem[{Beck et~al.(2024)Beck, Schuff, Lauscher, and
  Gurevych}]{beck2024sensitivity}
Tilman Beck, Hendrik Schuff, Anne Lauscher, and Iryna Gurevych. 2024.
\newblock \href {https://aclanthology.org/2024.eacl-long.159} {Sensitivity,
  performance, robustness: Deconstructing the effect of sociodemographic
  prompting}.
\newblock In \emph{Proceedings of the 18th Conference of the European Chapter
  of the Association for Computational Linguistics (Volume 1: Long Papers)},
  pages 2589--2615, St. Julian{'}s, Malta. Association for Computational
  Linguistics.

\bibitem[{Bonaldi et~al.(2024)Bonaldi, Damo, Ocampo, Cabrio, Villata, and
  Guerini}]{bonaldi2024safer}
Helena Bonaldi, Greta Damo, Nicol{\'a}s~Benjam{\'\i}n Ocampo, Elena Cabrio,
  Serena Villata, and Marco Guerini. 2024.
\newblock Is safer better? the impact of guardrails on the argumentative
  strength of llms in hate speech countering.
\newblock \emph{arXiv preprint arXiv:2410.03466}.

\bibitem[{Brown et~al.(2020)Brown, Mann, Ryder, Subbiah, Kaplan, Dhariwal,
  Neelakantan, Shyam, Sastry, Askell et~al.}]{brown2020language}
Tom Brown, Benjamin Mann, Nick Ryder, Melanie Subbiah, Jared~D Kaplan, Prafulla
  Dhariwal, Arvind Neelakantan, Pranav Shyam, Girish Sastry, Amanda Askell,
  et~al. 2020.
\newblock Language models are few-shot learners.
\newblock \emph{Advances in neural information processing systems},
  33:1877--1901.

\bibitem[{Choi et~al.(2023)Choi, Pei, Kumar, Shu, and Jurgens}]{choi2023llms}
Minje Choi, Jiaxin Pei, Sagar Kumar, Chang Shu, and David Jurgens. 2023.
\newblock Do llms understand social knowledge? evaluating the sociability of
  large language models with socket benchmark.
\newblock \emph{arXiv preprint arXiv:2305.14938}.

\bibitem[{Chung et~al.(2022)Chung, Hou, Longpre, Zoph, Tay, Fedus, Li, Wang,
  Dehghani, Brahma, Webson, Gu, Dai, Suzgun, Chen, Chowdhery, Castro-Ros,
  Pellat, Robinson, Valter, Narang, Mishra, Yu, Zhao, Huang, Dai, Yu, Petrov,
  Chi, Dean, Devlin, Roberts, Zhou, Le, and Wei}]{chung2022scaling}
Hyung~Won Chung, Le~Hou, Shayne Longpre, Barret Zoph, Yi~Tay, William Fedus,
  Yunxuan Li, Xuezhi Wang, Mostafa Dehghani, Siddhartha Brahma, Albert Webson,
  Shixiang~Shane Gu, Zhuyun Dai, Mirac Suzgun, Xinyun Chen, Aakanksha
  Chowdhery, Alex Castro-Ros, Marie Pellat, Kevin Robinson, Dasha Valter,
  Sharan Narang, Gaurav Mishra, Adams Yu, Vincent Zhao, Yanping Huang, Andrew
  Dai, Hongkun Yu, Slav Petrov, Ed~H. Chi, Jeff Dean, Jacob Devlin, Adam
  Roberts, Denny Zhou, Quoc~V. Le, and Jason Wei. 2022.
\newblock \href {http://arxiv.org/abs/2210.11416} {Scaling
  instruction-finetuned language models}.

\bibitem[{Deshpande et~al.(2023)Deshpande, Murahari, Rajpurohit, Kalyan, and
  Narasimhan}]{deshpande2023toxicity}
Ameet Deshpande, Vishvak Murahari, Tanmay Rajpurohit, Ashwin Kalyan, and
  Karthik Narasimhan. 2023.
\newblock \href {http://arxiv.org/abs/2304.05335} {Toxicity in chatgpt:
  Analyzing persona-assigned language models}.

\bibitem[{Dominguez-Olmedo et~al.(2023)Dominguez-Olmedo, Hardt, and
  Mendler-D{\"u}nner}]{dominguez2023questioning}
Ricardo Dominguez-Olmedo, Moritz Hardt, and Celestine Mendler-D{\"u}nner. 2023.
\newblock Questioning the survey responses of large language models.
\newblock \emph{arXiv preprint arXiv:2306.07951}.

\bibitem[{Dubey et~al.(2024)Dubey, Jauhri, Pandey, Kadian, Al-Dahle, Letman,
  Mathur, Schelten, Yang, Fan et~al.}]{dubey2024llama3}
Abhimanyu Dubey, Abhinav Jauhri, Abhinav Pandey, Abhishek Kadian, Ahmad
  Al-Dahle, Aiesha Letman, Akhil Mathur, Alan Schelten, Amy Yang, Angela Fan,
  et~al. 2024.
\newblock The llama 3 herd of models.
\newblock \emph{arXiv preprint arXiv:2407.21783}.

\bibitem[{Feng et~al.(2023)Feng, Park, Liu, and Tsvetkov}]{feng2023pretraining}
Shangbin Feng, Chan~Young Park, Yuhan Liu, and Yulia Tsvetkov. 2023.
\newblock \href {https://doi.org/10.18653/v1/2023.acl-long.656} {From
  pretraining data to language models to downstream tasks: Tracking the trails
  of political biases leading to unfair {NLP} models}.
\newblock In \emph{Proceedings of the 61st Annual Meeting of the Association
  for Computational Linguistics (Volume 1: Long Papers)}, pages 11737--11762,
  Toronto, Canada. Association for Computational Linguistics.

\bibitem[{Hancock(2007)}]{hancock2007multiplication}
Ange-Marie Hancock. 2007.
\newblock When multiplication doesn't equal quick addition: Examining
  intersectionality as a research paradigm.
\newblock \emph{Perspectives on politics}, 5(1):63--79.

\bibitem[{Hwang et~al.(2023)Hwang, Majumder, and Tandon}]{hwang2023aligning}
EunJeong Hwang, Bodhisattwa~Prasad Majumder, and Niket Tandon. 2023.
\newblock \href {http://arxiv.org/abs/2305.14929} {Aligning language models to
  user opinions}.

\bibitem[{Ivison et~al.(2023)Ivison, Wang, Pyatkin, Lambert, Peters, Dasigi,
  Jang, Wadden, Smith, Beltagy, and Hajishirzi}]{ivison2023camels}
Hamish Ivison, Yizhong Wang, Valentina Pyatkin, Nathan Lambert, Matthew Peters,
  Pradeep Dasigi, Joel Jang, David Wadden, Noah~A. Smith, Iz~Beltagy, and
  Hannaneh Hajishirzi. 2023.
\newblock \href {http://arxiv.org/abs/2311.10702} {Camels in a changing
  climate: Enhancing lm adaptation with tulu 2}.

\bibitem[{Jiang et~al.(2023)Jiang, Sablayrolles, Mensch, Bamford, Chaplot,
  de~las Casas, Bressand, Lengyel, Lample, Saulnier, Lavaud, Lachaux, Stock,
  Scao, Lavril, Wang, Lacroix, and Sayed}]{jiang2023mistral7b}
Albert~Q. Jiang, Alexandre Sablayrolles, Arthur Mensch, Chris Bamford,
  Devendra~Singh Chaplot, Diego de~las Casas, Florian Bressand, Gianna Lengyel,
  Guillaume Lample, Lucile Saulnier, Lélio~Renard Lavaud, Marie-Anne Lachaux,
  Pierre Stock, Teven~Le Scao, Thibaut Lavril, Thomas Wang, Timothée Lacroix,
  and William~El Sayed. 2023.
\newblock \href {http://arxiv.org/abs/2310.06825} {Mistral 7b}.

\bibitem[{Liang et~al.(2021)Liang, Wu, Morency, and
  Salakhutdinov}]{liang2021towards}
Paul~Pu Liang, Chiyu Wu, Louis-Philippe Morency, and Ruslan Salakhutdinov.
  2021.
\newblock Towards understanding and mitigating social biases in language
  models.
\newblock In \emph{International Conference on Machine Learning}, pages
  6565--6576. PMLR.

\bibitem[{Miehling et~al.(2024)Miehling, Desmond, Ramamurthy, Daly, Dognin,
  Rios, Bouneffouf, and Liu}]{miehling2024evaluating}
Erik Miehling, Michael Desmond, Karthikeyan~Natesan Ramamurthy, Elizabeth~M
  Daly, Pierre Dognin, Jesus Rios, Djallel Bouneffouf, and Miao Liu. 2024.
\newblock Evaluating the prompt steerability of large language models.
\newblock \emph{arXiv preprint arXiv:2411.12405}.

\bibitem[{Mu et~al.(2023)Mu, Wu, Thorne, Robinson, Aletras, Scarton, Bontcheva,
  and Song}]{mu2023navigating}
Yida Mu, Ben~P. Wu, William Thorne, Ambrose Robinson, Nikolaos Aletras,
  Carolina Scarton, Kalina Bontcheva, and Xingyi Song. 2023.
\newblock \href {http://arxiv.org/abs/2305.14310} {Navigating prompt complexity
  for zero-shot classification: A study of large language models in
  computational social science}.

\bibitem[{OpenAI(2023)}]{openai2023gpt4}
OpenAI. 2023.
\newblock \href {http://arxiv.org/abs/2303.08774} {Gpt-4 technical report}.

\bibitem[{Pei and Jurgens(2023)}]{pei2023annotator}
Jiaxin Pei and David Jurgens. 2023.
\newblock When do annotator demographics matter? measuring the influence of
  annotator demographics with the popquorn dataset.
\newblock In \emph{Proceedings of the 17th Linguistic Annotation Workshop
  (LAW-XVII) @ACL 2023}.

\bibitem[{Plaza-del arco et~al.(2023)Plaza-del arco, Nozza, and
  Hovy}]{plaza-del-arco2023-respectful}
Flor~Miriam Plaza-del arco, Debora Nozza, and Dirk Hovy. 2023.
\newblock \href {https://doi.org/10.18653/v1/2023.woah-1.6} {Respectful or
  toxic? using zero-shot learning with language models to detect hate speech}.
\newblock In \emph{The 7th Workshop on Online Abuse and Harms (WOAH)}, pages
  60--68, Toronto, Canada. Association for Computational Linguistics.

\bibitem[{Qwen et~al.(2025)Qwen, :, Yang, Yang, Zhang, Hui, Zheng, Yu, Li, Liu,
  Huang, Wei, Lin, Yang, Tu, Zhang, Yang, Yang, Zhou, Lin, Dang, Lu, Bao, Yang,
  Yu, Li, Xue, Zhang, Zhu, Men, Lin, Li, Tang, Xia, Ren, Ren, Fan, Su, Zhang,
  Wan, Liu, Cui, Zhang, and Qiu}]{qwen2025qwen25technicalreport}
Qwen, :, An~Yang, Baosong Yang, Beichen Zhang, Binyuan Hui, Bo~Zheng, Bowen Yu,
  Chengyuan Li, Dayiheng Liu, Fei Huang, Haoran Wei, Huan Lin, Jian Yang,
  Jianhong Tu, Jianwei Zhang, Jianxin Yang, Jiaxi Yang, Jingren Zhou, Junyang
  Lin, Kai Dang, Keming Lu, Keqin Bao, Kexin Yang, Le~Yu, Mei Li, Mingfeng Xue,
  Pei Zhang, Qin Zhu, Rui Men, Runji Lin, Tianhao Li, Tianyi Tang, Tingyu Xia,
  Xingzhang Ren, Xuancheng Ren, Yang Fan, Yang Su, Yichang Zhang, Yu~Wan,
  Yuqiong Liu, Zeyu Cui, Zhenru Zhang, and Zihan Qiu. 2025.
\newblock \href {http://arxiv.org/abs/2412.15115} {Qwen2.5 technical report}.

\bibitem[{Radford et~al.(2019)Radford, Wu, Child, Luan, Amodei, Sutskever
  et~al.}]{radford2019language}
Alec Radford, Jeffrey Wu, Rewon Child, David Luan, Dario Amodei, Ilya
  Sutskever, et~al. 2019.
\newblock Language models are unsupervised multitask learners.
\newblock \emph{OpenAI blog}, 1(8):9.

\bibitem[{Rytting et~al.(2023)Rytting, Sorensen, Argyle, Busby, Fulda, Gubler,
  and Wingate}]{rytting2023coding}
Christopher~Michael Rytting, Taylor Sorensen, Lisa Argyle, Ethan Busby, Nancy
  Fulda, Joshua Gubler, and David Wingate. 2023.
\newblock \href {http://arxiv.org/abs/2306.02177} {Towards coding social
  science datasets with language models}.

\bibitem[{Santurkar et~al.(2023{\natexlab{a}})Santurkar, Durmus, Ladhak, Lee,
  Liang, and Hashimoto}]{santurkar2023opinions}
Shibani Santurkar, Esin Durmus, Faisal Ladhak, Cinoo Lee, Percy Liang, and
  Tatsunori Hashimoto. 2023{\natexlab{a}}.
\newblock \href {http://arxiv.org/abs/2303.17548} {Whose opinions do language
  models reflect?}

\bibitem[{Santurkar et~al.(2023{\natexlab{b}})Santurkar, Durmus, Ladhak, Lee,
  Liang, and Hashimoto}]{santurkar2023whose}
Shibani Santurkar, Esin Durmus, Faisal Ladhak, Cinoo Lee, Percy Liang, and
  Tatsunori Hashimoto. 2023{\natexlab{b}}.
\newblock Whose opinions do language models reflect?
\newblock \emph{arXiv preprint arXiv:2303.17548}.

\bibitem[{Sorensen et~al.(2024)Sorensen, Moore, Fisher, Gordon, Mireshghallah,
  Rytting, Ye, Jiang, Lu, Dziri et~al.}]{sorensen2024roadmap}
Taylor Sorensen, Jared Moore, Jillian Fisher, Mitchell Gordon, Niloofar
  Mireshghallah, Christopher~Michael Rytting, Andre Ye, Liwei Jiang, Ximing Lu,
  Nouha Dziri, et~al. 2024.
\newblock A roadmap to pluralistic alignment.
\newblock \emph{arXiv preprint arXiv:2402.05070}.

\bibitem[{Tay et~al.(2023)Tay, Dehghani, Tran, Garcia, Wei, Wang, Chung,
  Shakeri, Bahri, Schuster, Zheng, Zhou, Houlsby, and Metzler}]{tay2023ul2}
Yi~Tay, Mostafa Dehghani, Vinh~Q. Tran, Xavier Garcia, Jason Wei, Xuezhi Wang,
  Hyung~Won Chung, Siamak Shakeri, Dara Bahri, Tal Schuster, Huaixiu~Steven
  Zheng, Denny Zhou, Neil Houlsby, and Donald Metzler. 2023.
\newblock \href {http://arxiv.org/abs/2205.05131} {Ul2: Unifying language
  learning paradigms}.

\bibitem[{Wang et~al.(2023)Wang, Peng, Que, Liu, Zhou, Wu, Guo, Gan, Ni, Zhang
  et~al.}]{wang2023rolellm}
Zekun~Moore Wang, Zhongyuan Peng, Haoran Que, Jiaheng Liu, Wangchunshu Zhou,
  Yuhan Wu, Hongcheng Guo, Ruitong Gan, Zehao Ni, Man Zhang, et~al. 2023.
\newblock Rolellm: Benchmarking, eliciting, and enhancing role-playing
  abilities of large language models.
\newblock \emph{arXiv preprint arXiv:2310.00746}.

\bibitem[{Yildirim and Paul(2024)}]{yildirim2024task}
Ilker Yildirim and LA~Paul. 2024.
\newblock From task structures to world models: what do llms know?
\newblock \emph{Trends in Cognitive Sciences}.

\bibitem[{Zhou et~al.(2023)Zhou, Zhu, Mathur, Zhang, Yu, Qi, Morency, Bisk,
  Fried, Neubig, and Sap}]{zhou2023sotopia}
Xuhui Zhou, Hao Zhu, Leena Mathur, Ruohong Zhang, Haofei Yu, Zhengyang Qi,
  Louis-Philippe Morency, Yonatan Bisk, Daniel Fried, Graham Neubig, and
  Maarten Sap. 2023.
\newblock \href {http://arxiv.org/abs/2310.11667} {Sotopia: Interactive
  evaluation for social intelligence in language agents}.

\bibitem[{Ziems et~al.(2023)Ziems, Held, Shaikh, Chen, Zhang, and
  Yang}]{ziems2023large}
Caleb Ziems, William Held, Omar Shaikh, Jiaao Chen, Zhehao Zhang, and Diyi
  Yang. 2023.
\newblock \href {http://arxiv.org/abs/2305.03514} {Can large language models
  transform computational social science?}

\end{thebibliography}

\clearpage

\appendix

\section*{Appendix}
\label{sec:appendix}

\section{Data}
\label{appendix:data}

\begin{figure*}[t!]
\centering
    \includegraphics[width=1.0\textwidth]{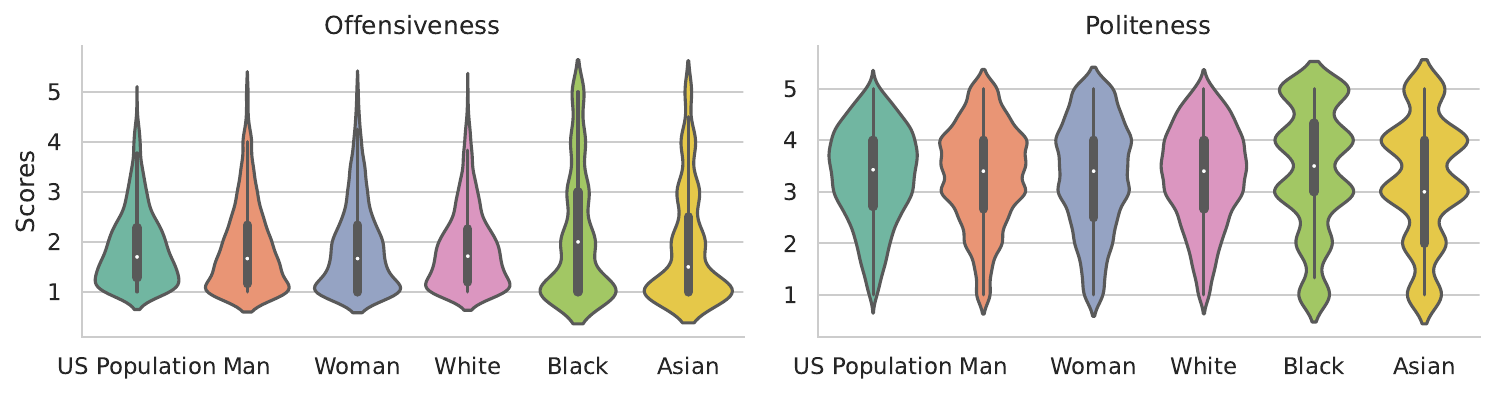}
    \caption{Distribution of annotations from different demographic groups for both offensiveness and politeness tasks.}
    \label{fig:score}
\end{figure*}

Table~\ref{tab:desc_offensive} and Table~\ref{tab:desc_polite} demonstrate the descriptives for offensiveness and politeness ratings. Figure~\ref{fig:score} further visualizes the distributions of annotations by demographic groups for both tasks.

\begin{table}[h!]
\centering
\resizebox{0.48\textwidth}{!}{ 
\begin{tabular}{lllllll}
\hline
     & Overall & Man   & Woman & White & Black & Asian \\ \hline
Mean & 1.881   & 1.880 & 1.893 & 1.865 & 2.096 & 1.906 \\
Std  & 0.756   & 0.861 & 0.893 & 0.790 & 1.270 & 1.147 \\
N    & 1500    & 1483  & 1488  & 1500  & 1003  & 708   \\ \hline
\end{tabular}
}
\caption{Descriptives of offensiveness ratings by demographic groups.}
\label{tab:desc_offensive}
\end{table}

\begin{table}[h!]
\centering
\resizebox{0.48\textwidth}{!}{ 
\begin{tabular}{lllllll}
\hline
     & Overall & Man   & Woman & White & Black & Asian \\ \hline
Mean & 3.305   & 3.318 & 3.308 & 3.304 & 3.403 & 3.196 \\
Std  & 0.910   & 0.956 & 1.070 & 0.946 & 1.227 & 1.182 \\
N    & 3718    & 3660  & 3681  & 3717  & 2222  & 1327  \\ \hline
\end{tabular}
}
\caption{Descriptives of politeness ratings by demographic groups.}
\label{tab:desc_polite}
\end{table}

\section{Prompts}
\label{appendix:prompt}
Table~\ref{tab:prompt} provides an example of the prompts used in the experiments.

\begin{table}[h]
\centering

\scalebox{0.95}{
\begin{tikzpicture}
\fontsize{9}{11}\selectfont
\node[draw, inner sep=5pt, rounded corners, fill=gray!20]{
    \begin{tabular}{p{65mm}}

    \texttt{How does a person rate how offensive the sentence is from 1 (not offensive at all) to 5 (very offensive)} 
    
    \texttt{Options: 1, 2, 3, 4, 5}

    \texttt{\{Text\}}
    
    \texttt{Response (Provide the answer without explaining your reasoning):}
\end{tabular}
};
\end{tikzpicture}}

\caption{An example prompt for this study.}
\label{tab:prompt}
\end{table}

\paragraph{Robustness check} 

\begin{figure*}[t!]
    \centering
    \includegraphics[width=1.0\textwidth]{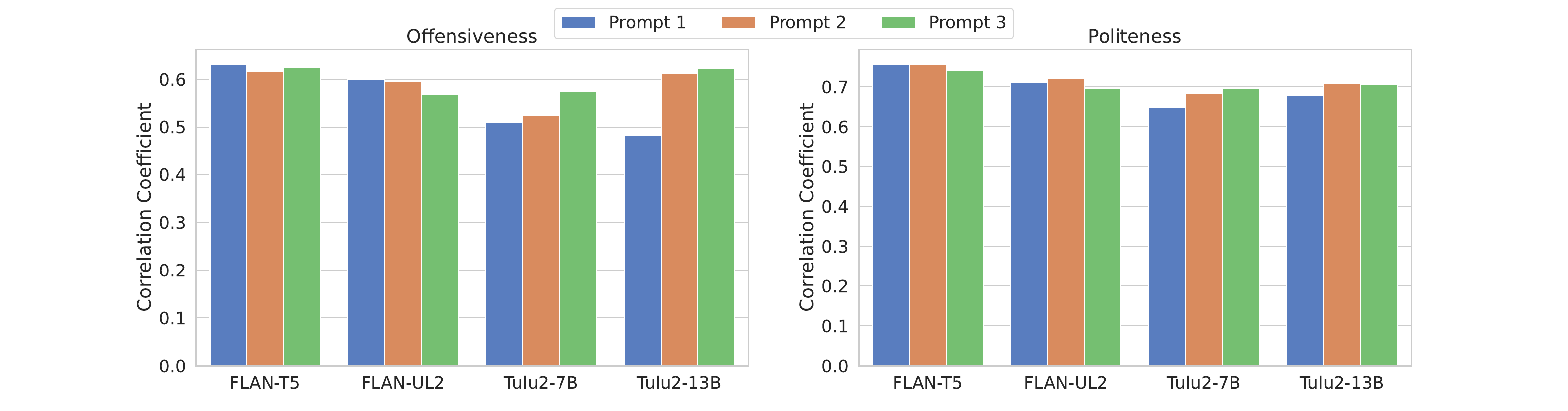} 
    \vspace{-0.5cm}
    \caption{Models' performances do not change a lot when being prompted with different templates.}
    \label{fig:robust}
\end{figure*}

We test the robustness of our results with different prompt templates on four open-source LLMs: FLAN-T5, FLAN-UL2, Tulu2-DPO-7B, and Tulu2-DPO-13B. In the robustness check, we calculate the correlation coefficient between the LLMs’ baseline predictions and the overall annotations representing the US population (the aggregated ratings for the entire sample). This overall correlation coefficient serves as a reference point for comparing the effects of applying different prompt templates. As shown in Figure \ref{fig:robust}, prompt templates have limited influence on the correlation coefficients between base prompts without identity words and labels representing the U.S population. It indicates that models' perception of offensivenss and politeness does not change much with different ways of prompting.

Table~\ref{tab:prompt_list} lists the three prompt templates that are used as a robustness check.

\begin{table}[h]
\resizebox{\columnwidth}{!}{%
\begin{tabular}{|l|p{8cm}|}
\hline
  & Templates                                                                          \\ \hline
1 & How does \{identity\} rate how \{offensive/polite\} the sentence is from 1 (not \{offensive/polite\} at all) to 5 (very \{offensive/polite\}). Options: 1, 2, 3, 4, 5: \{text\}"                                             \\ \hline
2 & If you are \{identity\}, how would you rate the \{offensiveness/politeness\} of the given sentence from 1 (not \{offensive/polite\} at all) to 5 (very \{offensive/polite\}). Options: 1, 2, 3, 4, 5: \{text\}"              \\ \hline
3 & Rate the \{offensiveness/politeness\} of the following sentence using a scale of 1 (not \{offensive/polite\} at all) to 5 (very \{offensive/polite\}) from the perspective of \{identity\}. Options: 1, 2, 3, 4, 5: \{text\} \\ \hline
\end{tabular}%
}
\caption{Prompt templates used in our study}
\label{tab:prompt_list}
\end{table}

\section{Experimentation details}
\label{appendix:method}

\paragraph{Model Settings} We generally follow the default settings recommended by the model developers.  For model temperature, we set it to 0 for a more deterministic generation approach. This setting allows for a better understanding of the question “what is the most likely answer a model will return given a question and a specific group association?”

\paragraph{Computing infrastructure} We use three NVIDIA RTX A5000 GPUs in our experiment. A full run of our experiment takes around 6 hours of GPU time.

\paragraph{Data and model license} We have checked that our usage of the data and models are in compliance with the corresponding license.

\paragraph{Packages} We used the following packages in our experiment: \texttt{accelarate}, \texttt{datasets}, \texttt{numpy}, \texttt{pandas}, \texttt{seaborn}, \texttt{statsmodels}, \texttt{transformers}.

\section{Regression Results}
\label{appendix:results}
 
In Table~\ref{tab:basegap} and Table~\ref{tab:addgap}, we report the estimated fixed effects of predictors, along with their standard errors and statistical significance. Statistical significance is denoted by stars, where a p-value less than 0.05 is marked with one star (*), a p-value less than 0.01 is marked with two stars (**), and a p-value less than 0.001 is marked with three stars (***).

\begin{table*}[t!]
\centering
\resizebox{\textwidth}{!}{%
\begin{tabular}{lccccccccc}
\hline
\multicolumn{1}{c}{}      & FLAN-T5                                                     & FLAN-UL2                                                   & Tulu2-7B                                                   & Tulu2-13B                                                  & GPT3.5                                                     & GPT4                                                        & Llama3.1-8B                                                 & Mistral0.3-7B                                              & Qwen2.5-7B                                                 \\ \hline
\multicolumn{10}{l}{\textit{\textbf{Offensiveness, Gender (reference=Man)}}}                                                                                                                                                                                                                                                                                                                                                                                                                                                                                                                      \\
\multicolumn{1}{c}{Woman} & \begin{tabular}[c]{@{}c@{}}-0.034\\ (0.017)\end{tabular}    & \begin{tabular}[c]{@{}c@{}}-0.046*\\ (0.019)\end{tabular}  & \begin{tabular}[c]{@{}c@{}}-0.024\\ (0.021)\end{tabular}   & \begin{tabular}[c]{@{}c@{}}0.006\\ (0.019)\end{tabular}    & \begin{tabular}[c]{@{}c@{}}-0.015\\ (0.016)\end{tabular}   & \begin{tabular}[c]{@{}c@{}}-0.036\\ (0.019)\end{tabular}    & \begin{tabular}[c]{@{}c@{}}-0.022\\ (0.019)\end{tabular}    & \begin{tabular}[c]{@{}c@{}}-0.027\\ (0.02)\end{tabular}    & \begin{tabular}[c]{@{}c@{}}-0.024\\ (0.019)\end{tabular}   \\ \hline
\multicolumn{10}{l}{\textit{\textbf{Offensiveness, Ethnicity (reference=White)}}}                                                                                                                                                                                                                                                                                                                                                                                                                                                                                                                 \\
Black                     & \begin{tabular}[c]{@{}c@{}}0.231***\\ (0.027)\end{tabular}  & \begin{tabular}[c]{@{}c@{}}0.064*\\ (0.031)\end{tabular}   & \begin{tabular}[c]{@{}c@{}}-0.068*\\ (0.033)\end{tabular}  & \begin{tabular}[c]{@{}c@{}}0.056*\\ (0.027)\end{tabular}   & \begin{tabular}[c]{@{}c@{}}0.319***\\ (0.023)\end{tabular} & \begin{tabular}[c]{@{}c@{}}0.222***\\ (0.031)\end{tabular}  & \begin{tabular}[c]{@{}c@{}}0.085**\\ (0.03)\end{tabular}    & \begin{tabular}[c]{@{}c@{}}0.041\\ (0.032)\end{tabular}    & \begin{tabular}[c]{@{}c@{}}0.038\\ (0.03)\end{tabular}     \\
Asian                     & \begin{tabular}[c]{@{}c@{}}0.252***\\ (0.031)\end{tabular}  & \begin{tabular}[c]{@{}c@{}}0.049\\ (0.035)\end{tabular}    & \begin{tabular}[c]{@{}c@{}}0.016\\ (0.038)\end{tabular}    & \begin{tabular}[c]{@{}c@{}}0.127***\\ (0.031)\end{tabular} & \begin{tabular}[c]{@{}c@{}}0.267***\\ (0.027)\end{tabular} & \begin{tabular}[c]{@{}c@{}}0.219***\\ (0.035)\end{tabular}  & \begin{tabular}[c]{@{}c@{}}0.088**\\ (0.034)\end{tabular}   & \begin{tabular}[c]{@{}c@{}}0.079*\\ (0.037)\end{tabular}   & \begin{tabular}[c]{@{}c@{}}0.113**\\ (0.034)\end{tabular}  \\ \hline
\multicolumn{10}{l}{\textit{\textbf{Politeness, Gender (reference=Man)}}}                                                                                                                                                                                                                                                                                                                                                                                                                                                                                                                         \\
\multicolumn{1}{c}{Woman} & \begin{tabular}[c]{@{}c@{}}-0.059***\\ (0.012)\end{tabular} & \begin{tabular}[c]{@{}c@{}}-0.04**\\ (0.013)\end{tabular}  & \begin{tabular}[c]{@{}c@{}}-0.002\\ (0.08)\end{tabular}    & \begin{tabular}[c]{@{}c@{}}-0.023*\\ (0.011)\end{tabular}  & \begin{tabular}[c]{@{}c@{}}-0.008\\ (0.011)\end{tabular}   & \begin{tabular}[c]{@{}c@{}}-0.065***\\ (0.012)\end{tabular} & \begin{tabular}[c]{@{}c@{}}-0.047***\\ (0.012)\end{tabular} & \begin{tabular}[c]{@{}c@{}}-0.007\\ (0.011)\end{tabular}   & \begin{tabular}[c]{@{}c@{}}-0.02\\ (0.012)\end{tabular}    \\ \hline
\multicolumn{10}{l}{\textit{\textbf{Politeness, Ethnicity (reference=White)}}}                                                                                                                                                                                                                                                                                                                                                                                                                                                                                                                    \\
Black                     & \begin{tabular}[c]{@{}c@{}}0.158***\\ (0.017)\end{tabular}  & \begin{tabular}[c]{@{}c@{}}0.068***\\ (0.019)\end{tabular} & \begin{tabular}[c]{@{}c@{}}0.218***\\ (0.017)\end{tabular} & \begin{tabular}[c]{@{}c@{}}0.238***\\ (0.015)\end{tabular} & \begin{tabular}[c]{@{}c@{}}0.27***\\ (0.015)\end{tabular}  & \begin{tabular}[c]{@{}c@{}}0.106***\\ (0.017)\end{tabular}  & \begin{tabular}[c]{@{}c@{}}0.187***\\ (0.017)\end{tabular}  & \begin{tabular}[c]{@{}c@{}}0.241***\\ (0.016)\end{tabular} & \begin{tabular}[c]{@{}c@{}}0.231***\\ (0.017)\end{tabular} \\
Asian                     & \begin{tabular}[c]{@{}c@{}}0.135***\\ (0.021)\end{tabular}  & \begin{tabular}[c]{@{}c@{}}0.14***\\ (0.023)\end{tabular}  & \begin{tabular}[c]{@{}c@{}}0.2***\\ (0.02)\end{tabular}    & \begin{tabular}[c]{@{}c@{}}0.204***\\ (0.019)\end{tabular} & \begin{tabular}[c]{@{}c@{}}0.206***\\ (0.018)\end{tabular} & \begin{tabular}[c]{@{}c@{}}0.172***\\ (0.021)\end{tabular}  & \begin{tabular}[c]{@{}c@{}}0.132***\\ (0.021)\end{tabular}  & \begin{tabular}[c]{@{}c@{}}0.177***\\ (0.02)\end{tabular}  & \begin{tabular}[c]{@{}c@{}}0.107***\\ (0.02)\end{tabular}  \\ \hline
\end{tabular}
}

\caption{Regression results for predicting the gap between zero-shot model predictions and the labels from each demographic group. }
\label{tab:basegap}
\end{table*}

\begin{table*}[t!]
\centering
\resizebox{\textwidth}{!}{%
\begin{tabular}{lccccccccc}
\hline
\multicolumn{1}{c}{}      & FLAN-T5                                                     & FLAN-UL2                                                    & Tulu2-7B                                                   & Tulu2-13B                                                   & GPT3.5                                                      & GPT4                                                        & Llama3.1-8B                                                & Mistral0.3-7B                                               & Qwen2.5-7B                                                  \\ \hline
\multicolumn{10}{l}{\textit{\textbf{Offensiveness, Gender}}}                                                                                                                                                                                                                                                                                                                                                                                                                                                                                                                                          \\
Man                       & \begin{tabular}[c]{@{}c@{}}0.054***\\ (0.009)\end{tabular}  & \begin{tabular}[c]{@{}c@{}}0.056***\\ (0.013)\end{tabular}  & \begin{tabular}[c]{@{}c@{}}-0.06***\\ (0.015)\end{tabular} & \begin{tabular}[c]{@{}c@{}}-0.176***\\ (0.012)\end{tabular} & \begin{tabular}[c]{@{}c@{}}-0.051***\\ (0.011)\end{tabular} & \begin{tabular}[c]{@{}c@{}}-0.074***\\ (0.018)\end{tabular} & \begin{tabular}[c]{@{}c@{}}0.035\\ (0.019)\end{tabular}    & \begin{tabular}[c]{@{}c@{}}-0.065***\\ (0.011)\end{tabular} & \begin{tabular}[c]{@{}c@{}}-0.01\\ (0.014)\end{tabular}     \\
\multicolumn{1}{c}{Woman} & \begin{tabular}[c]{@{}c@{}}0.076***\\ (0.011)\end{tabular}  & \begin{tabular}[c]{@{}c@{}}0.073***\\ (0.014)\end{tabular}  & \begin{tabular}[c]{@{}c@{}}-0.04**\\ (0.015)\end{tabular}  & \begin{tabular}[c]{@{}c@{}}-0.265***\\ (0.014)\end{tabular} & \begin{tabular}[c]{@{}c@{}}-0.044**\\ (0.013)\end{tabular}  & \begin{tabular}[c]{@{}c@{}}-0.024\\ (0.017)\end{tabular}    & \begin{tabular}[c]{@{}c@{}}-0.057**\\ (0.019)\end{tabular} & \begin{tabular}[c]{@{}c@{}}0.104***\\ (0.014)\end{tabular}  & \begin{tabular}[c]{@{}c@{}}0.069***\\ (0.014)\end{tabular}  \\ \hline
\multicolumn{10}{l}{\textit{\textbf{Offensiveness, Ethnicity}}}                                                                                                                                                                                                                                                                                                                                                                                                                                                                                                                                       \\
White                     & \begin{tabular}[c]{@{}c@{}}-0.064***\\ (0.014)\end{tabular} & \begin{tabular}[c]{@{}c@{}}-0.09***\\ (0.016)\end{tabular}  & \begin{tabular}[c]{@{}c@{}}-0.16***\\ (0.018)\end{tabular} & \begin{tabular}[c]{@{}c@{}}-0.152***\\ (0.013)\end{tabular} & \begin{tabular}[c]{@{}c@{}}-0.097***\\ (0.016)\end{tabular} & \begin{tabular}[c]{@{}c@{}}0.01\\ (0.02)\end{tabular}       & \begin{tabular}[c]{@{}c@{}}-0.059**\\ (0.021)\end{tabular} & \begin{tabular}[c]{@{}c@{}}-0.232***\\ (0.017)\end{tabular} & \begin{tabular}[c]{@{}c@{}}-0.062***\\ (0.016)\end{tabular} \\
Black                     & \begin{tabular}[c]{@{}c@{}}0.033\\ (0.021)\end{tabular}     & \begin{tabular}[c]{@{}c@{}}-0.13***\\ (0.035)\end{tabular}  & \begin{tabular}[c]{@{}c@{}}-0.073**\\ (0.025)\end{tabular} & \begin{tabular}[c]{@{}c@{}}-0.015\\ (0.022)\end{tabular}    & \begin{tabular}[c]{@{}c@{}}-0.177***\\ (0.035)\end{tabular} & \begin{tabular}[c]{@{}c@{}}0.062\\ (0.037)\end{tabular}     & \begin{tabular}[c]{@{}c@{}}0.061*\\ (0.03)\end{tabular}    & \begin{tabular}[c]{@{}c@{}}0.311***\\ (0.03)\end{tabular}   & \begin{tabular}[c]{@{}c@{}}0.266***\\ (0.026)\end{tabular}  \\
Asian                     & \begin{tabular}[c]{@{}c@{}}0.008\\ (0.017)\end{tabular}     & \begin{tabular}[c]{@{}c@{}}-0.298***\\ (0.048)\end{tabular} & \begin{tabular}[c]{@{}c@{}}-0.078**\\ (0.028)\end{tabular} & \begin{tabular}[c]{@{}c@{}}-0.182***\\ (0.025)\end{tabular} & \begin{tabular}[c]{@{}c@{}}-0.108***\\ (0.029)\end{tabular} & \begin{tabular}[c]{@{}c@{}}-0.097*\\ (0.041\end{tabular}    & \begin{tabular}[c]{@{}c@{}}0.004\\ (0.035)\end{tabular}    & \begin{tabular}[c]{@{}c@{}}0.042\\ (0.027)\end{tabular}     & \begin{tabular}[c]{@{}c@{}}0.11***\\ (0.024)\end{tabular}   \\ \hline
\multicolumn{10}{l}{\textit{\textbf{Politeness, Gender}}}                                                                                                                                                                                                                                                                                                                                                                                                                                                                                                                                             \\
Man                       & \begin{tabular}[c]{@{}c@{}}0.031***\\ (0.005)\end{tabular}  & \begin{tabular}[c]{@{}c@{}}0.032***\\ (0.006)\end{tabular}  & \begin{tabular}[c]{@{}c@{}}-0.023**\\ (0.007)\end{tabular} & \begin{tabular}[c]{@{}c@{}}-0.007\\ (0.007)\end{tabular}    & \begin{tabular}[c]{@{}c@{}}-0.008\\ (0.005)\end{tabular}    & \begin{tabular}[c]{@{}c@{}}0.031***\\ (0.005)\end{tabular}  & \begin{tabular}[c]{@{}c@{}}0.001\\ (0.01)\end{tabular}     & \begin{tabular}[c]{@{}c@{}}-0.009\\ (0.005)\end{tabular}    & \begin{tabular}[c]{@{}c@{}}-0.013**\\ (0.006)\end{tabular}  \\
\multicolumn{1}{c}{Woman} & \begin{tabular}[c]{@{}c@{}}0.02***\\ (0.005)\end{tabular}   & \begin{tabular}[c]{@{}c@{}}-0.008\\ (0.006)\end{tabular}    & \begin{tabular}[c]{@{}c@{}}-0.01\\ (0.007)\end{tabular}    & \begin{tabular}[c]{@{}c@{}}0.026**\\ (0.008)\end{tabular}   & \begin{tabular}[c]{@{}c@{}}0.008\\ (0.004)\end{tabular}     & \begin{tabular}[c]{@{}c@{}}-0.016**\\ (0.005)\end{tabular}  & \begin{tabular}[c]{@{}c@{}}0.018\\ (0.01)\end{tabular}     & \begin{tabular}[c]{@{}c@{}}0.005\\ (0.006)\end{tabular}     & \begin{tabular}[c]{@{}c@{}}0.063***\\ (0.007)\end{tabular}  \\ \hline
\multicolumn{10}{l}{\textit{\textbf{Politeness, Ethnicity}}}                                                                                                                                                                                                                                                                                                                                                                                                                                                                                                                                          \\
White                     & \begin{tabular}[c]{@{}c@{}}0.04***\\ (0.005)\end{tabular}   & \begin{tabular}[c]{@{}c@{}}-0.007\\ (0.005)\end{tabular}    & \begin{tabular}[c]{@{}c@{}}-0.007\\ (0.008)\end{tabular}   & \begin{tabular}[c]{@{}c@{}}0.046***\\ (0.007)\end{tabular}  & \begin{tabular}[c]{@{}c@{}}0.019***\\ (0.006)\end{tabular}  & \begin{tabular}[c]{@{}c@{}}0.039***\\ (0.005)\end{tabular}  & \begin{tabular}[c]{@{}c@{}}0.009\\ (0.01)\end{tabular}     & \begin{tabular}[c]{@{}c@{}}0.005\\ (0.006)\end{tabular}     & \begin{tabular}[c]{@{}c@{}}-0.006\\ (0.007)\end{tabular}    \\
Black                     & \begin{tabular}[c]{@{}c@{}}0.04***\\ (0.009)\end{tabular}   & \begin{tabular}[c]{@{}c@{}}-0.034**\\ (0.01)\end{tabular}   & \begin{tabular}[c]{@{}c@{}}-0.035**\\ (0.012)\end{tabular} & \begin{tabular}[c]{@{}c@{}}0.014\\ (0.014)\end{tabular}     & \begin{tabular}[c]{@{}c@{}}0.034**\\ (0.012)\end{tabular}   & \begin{tabular}[c]{@{}c@{}}0.128***\\ (0.012)\end{tabular}  & \begin{tabular}[c]{@{}c@{}}0.017\\ (0.015)\end{tabular}    & \begin{tabular}[c]{@{}c@{}}0.035**\\ (0.012)\end{tabular}   & \begin{tabular}[c]{@{}c@{}}0.154***\\ (0.013)\end{tabular}  \\
Asian                     & \begin{tabular}[c]{@{}c@{}}-0.013\\ (0.012)\end{tabular}    & \begin{tabular}[c]{@{}c@{}}-0.092***\\ (0.015)\end{tabular} & \begin{tabular}[c]{@{}c@{}}-0.0\\ (0.015)\end{tabular}     & \begin{tabular}[c]{@{}c@{}}-0.021\\ (0.015)\end{tabular}    & \begin{tabular}[c]{@{}c@{}}0.048**\\ (0.015)\end{tabular}   & \begin{tabular}[c]{@{}c@{}}-0.107***\\ (0.012)\end{tabular} & \begin{tabular}[c]{@{}c@{}}0.079***\\ (0.02)\end{tabular}  & \begin{tabular}[c]{@{}c@{}}0.044**\\ (0.014)\end{tabular}   & \begin{tabular}[c]{@{}c@{}}0.113***\\ (0.014)\end{tabular}  \\ \hline
\end{tabular}
}

\caption{Regression results for predicting the prediction errors when adding identity to the prompt, relative to an identity-free prompt.}
\label{tab:addgap}
\end{table*}

\section{LLM Guardrails}
\label{appendix:guardrail}

When responding to potentially harmful queries, LLMs may refuse to provide an answer due to implicit guardrails designed to mitigate biases and protect users from inappropriate content. Table~\ref{tab:invalid_offensive} and Table~\ref{tab:invalid_polite} summarize the percentages of invalid responses across nine LLMs when prompted with and without specific demographic information.

\begin{table*}[t!]
\centering
\begin{tabular}{llllllll}
\hline
              &  & Base  & Man   & Woman  & White  & Black  & Asian  \\ \hline
FLAN-T5       &  & 0.0\% & 0.1\% & 0.1\%  & 0.1\%  & 0.1\%  & 0.0\%  \\
FLAN-UL2      &  & 0.0\% & 0.0\% & 0.0\%  & 0.0\%  & 0.2\%  & 0.1\%  \\
Tulu2-7B      &  & 7.5\% & 2.2\% & 3.6\%  & 6.3\%  & 14.3\% & 15.3\% \\
Tulu2-13B     &  & 2.0\% & 3.2\% & 3.2\%  & 3.4\%  & 20.0\% & 13.0\% \\
GPT 3.5       &  & 1.3\% & 4.0\% & 16.9\% & 23.1\% & 71.1\% & 44.8\% \\
GPT 4         &  & 0.0\% & 0.0\% & 0.0\%  & 0.0\%  & 0.0\%  & 0.0\%  \\
Llama3.1-8B   &  & 0.4\% & 0.6\% & 0.5\%  & 0.9\%  & 0.9\%  & 0.7\%  \\
Mistral0.3-7B &  & 4.3\% & 3.5\% & 3.9\%  & 13.5\% & 24.3\% & 13.7\% \\
Qwen2.5-7B    &  & 0.7\% & 0.6\% & 0.7\%  & 0.9\%  & 1.3\%   & 1.0\%  \\ \hline
\end{tabular}

\caption{Percentages of invalid responses on offensiveness task}
\label{tab:invalid_offensive}
\end{table*}

\begin{table*}[t!]
\centering
\begin{tabular}{llllllll}
\hline
              &  & Base  & Man   & Woman & White & Black  & Asian \\ \hline
FLAN-T5       &  & 0.0\% & 0.0\% & 0.0\% & 0.0\% & 0.0\%  & 0.0\% \\
FLAN-UL2      &  & 0.0\% & 0.1\% & 0.1\% & 0.1\% & 0.1\%  & 0.1\% \\
Tulu2-7B      &  & 2.8\% & 1.6\% & 2.7\% & 2.9\% & 13.1\% & 7.2\% \\
Tulu2-13B     &  & 1.7\% & 2.6\% & 2.6\% & 3.3\% & 9.7\%  & 4.0\% \\
GPT 3.5       &  & 0.1\% & 0.1\% & 0.1\% & 0.3\% & 6.5\%  & 0.2\% \\
GPT 4         &  & 0.0\% & 0.0\% & 0.0\% & 0.0\% & 0.0\%  & 0.0\% \\
Llama3.1-8B   &  & 0.0\% & 0.0\% & 0.1\% & 0.1\% & 0.2\%  & 0.2\% \\
Mistral0.3-7B &  & 0.4\% & 0.5\% & 0.9\% & 0.9\% & 3.6\%  & 1.3\% \\
Qwen2.5-7B    &  & 0.3\% & 0.4\% & 0.6\% & 0.6\% & 0.7\%  & 0.7\% \\ \hline
\end{tabular}

\caption{Percentages of invalid responses on politeness task}
\label{tab:invalid_polite}
\end{table*}

\section{Usage of AI Assistants}
We use AI assistants to check the grammar of our paper.

\end{document}